\pdfoutput=1

\documentclass[11pt]{article}

\usepackage[final]{acl}

\usepackage{times}
\usepackage{latexsym}

\usepackage[T1]{fontenc}

\usepackage[utf8]{inputenc}

\usepackage{microtype}

\usepackage{inconsolata}

\usepackage{graphicx}

\usepackage{booktabs}
\usepackage{siunitx}
\usepackage{xparse}
\usepackage{multirow}
\usepackage{tikz}
\usetikzlibrary{positioning,shapes,arrows.meta}
\graphicspath{{./}}

\title{Beyond the Numbers: Transparency in Relation Extraction Benchmark Creation and Leaderboards}

\author{
  Varvara Arzt \\
  Faculty of Informatics, TU Wien \\
  D!ARC, University of Klagenfurt \\
  \texttt{varvara.arzt@tuwien.ac.at} \\\And
  Allan Hanbury \\
  Faculty of Informatics, TU Wien \\
  \texttt{allan.hanbury@tuwien.ac.at} \\
}

\begin{document}
\maketitle
\begin{abstract}

This paper investigates the transparency in the creation of benchmarks and the use of leaderboards for measuring progress in NLP, with a focus on the relation extraction (RE) task. Existing RE benchmarks often suffer from insufficient documentation, lacking crucial details such as data sources, inter-annotator agreement, the algorithms used for the selection of instances for datasets, and information on potential biases like dataset imbalance. Progress in RE is frequently measured by leaderboards that rank systems based on evaluation methods, typically limited to aggregate metrics like F1-score. However, the absence of detailed performance analysis beyond these metrics can obscure the true generalisation capabilities of models. Our analysis reveals that widely used RE benchmarks, such as TACRED and NYT, tend to be highly imbalanced and contain noisy labels. Moreover, the lack of class-based performance metrics fails to accurately reflect model performance across datasets with a large number of relation types. These limitations should be carefully considered when reporting progress in RE. While our discussion centers on the transparency of RE benchmarks and leaderboards, the observations we discuss are broadly applicable to other NLP tasks as well. Rather than undermining the significance and value of existing RE benchmarks and the development of new models, this paper advocates for improved documentation and more rigorous evaluation to advance the field.
\end{abstract}

\section{Introduction}

We examine the transparency in benchmarks and leaderboards, focusing on the relation extraction (RE) task. Our analysis utilises two broadly accepted RE datasets, TACRED \citep{zhang-etal-2017-position} and NYT \citep{10.1007/978-3-642-15939-8_10}. While this paper focuses on the transparency of RE benchmarks and leaderboards, the observations we discuss are also relevant to other areas of natural language processing (NLP).

The development of state-of-the-art (SOTA) models in NLP is heavily reliant on benchmarks for evaluation. These benchmarks not only serve as a standard for assessing model performance but also play a pivotal role in shaping the perceived progress within the field. However, the current benchmarks often lack transparency in regard to their creation process, which can significantly impact the reliability of the evaluations conducted using them.

Opaque benchmarks and the absence of detailed performance analysis can obscure the true generalisation capabilities of models \citep{10.1145/3458723,dehghani2021benchmarklottery}. When benchmarks are not fully transparent — lacking comprehensive metadata, clear articulation of limitations, and rigorous evaluation reports — their ability to accurately reflect a model's robustness and generalisability is compromised. This can lead to an overestimation of progress, as models may appear to perform well on certain benchmarks but fail to generalise effectively to different or more complex datasets.

To enhance transparency and reproducibility in the evaluation of models, it is essential to publish the annotation guidelines and instructions that were provided to annotators during the creation of benchmarks. Understanding the exact criteria and procedures used in annotation is critical for interpreting the results obtained from these benchmarks and for comparing the performance of different models.

It is also important to recognise that widely used benchmarks such as TACRED \citep{zhang-etal-2017-position}, TACRED-RE \citep{alt-etal-2020-tacred}, and NYT \citep{10.1007/978-3-642-15939-8_10} cover only a subset of all possible relations. This limitation should be considered when evaluating models, as these benchmarks do not necessarily capture the full complexity of relation extraction task.

Additionally, when asserting that a new system outperforms existing ones, it is crucial to provide more granular results beyond aggregate metrics like weighted average or macro F1-score. These metrics, while useful, may not be sufficiently informative, particularly in the context of datasets with a large number of labels \citep{dehghani2021benchmarklottery} and significant class imbalances, such as NYT or TACRED.

Although this position paper addresses certain issues with existing RE benchmarks and evaluation approaches, it does not seek to diminish their significance or the value of developing new RE models, which are crucial for advancing the NLP field. Instead, it aims to promote improved documentation of benchmarks and the adoption of more rigorous evaluation practices for SOTA RE systems. 






\section{Related Work}

Despite the critical role of data in NLP, the documentation of the creation process of existing datasets remains scarce, unstandardised, and often lacks transparency, even for publicly available datasets \citep{bender-friedman-2018-data,10.1145/3458723,peng2021mitigating,SINGH2023144,kovatchev-lease-2024-benchmark}.

\citealp{10.1145/3458723} addresses the issue of insufficient transparency in dataset creation by proposing that dataset creators accompany each dataset with a datasheet. This datasheet would document essential information about the dataset's creation process, thereby enhancing the reproducibility of machine learning experiments and helping to mitigate potential biases. They outline seven key stages of the dataset lifecycle: motivation, composition, collection process, preprocessing/cleaning/labeling, intended uses, distribution, and maintenance.


The lack of transparency in benchmark creation significantly impacts the evaluation of models trained on these benchmarks. As \citealp{kovatchev-lease-2024-benchmark} highlights, many evaluation frameworks operate under the implicit assumption that a particular dataset is representative of the task it is intended to benchmark. However, systematic approaches to testing model generalisation remain limited \citep{hupkes2023taxonomy}. To address this gap, \citealp{kovatchev-lease-2024-benchmark} propose the use of dataset similarity vectors, which consider various dimensions of the data, such as noise and ambiguity features, to more accurately predict the generalisation capabilities of models trained on these datasets.

\citealp{hupkes2023taxonomy} present a comprehensive taxonomy of methods for studying the generalisation capabilities of models and introduce the GenBench evaluation card template\footnote{Available at \url{https://genbench.org/eval_cards/}.} to assist researchers in systematically documenting, justifying, and tracing their generalisation experiments. Evaluating the generalisation capabilities of models has become increasingly complex in the era of large language models (LLMs), which strive to achieve human-like generalisation but are trained on vast, uncontrolled, and often nontransparent datasets. 

To enhance the transparency of model evaluation processes, researchers advocate for testing new SOTA models in challenging scenarios involving perturbed instances \citep{wu-etal-2019-errudite,gardner-etal-2020-evaluating,goel-etal-2021-robustness}, thereby assessing model capabilities in more realistic settings than those provided by traditional test sets. \citealp{linzen-2020-accelerate}, when discussing the limitations of current evaluation approaches, particularly in the context of developing systems with human-like generalisation capabilities, introduces the Generalisation Leaderboards. These leaderboards evaluate systems on test sets derived from distributions different from those used during training. This approach addresses the limitation that testing a model on data drawn from the same distribution as the training set does not necessarily demonstrate the model's ability to effectively solve the task; rather, it may merely reflect the model's proficiency in capturing statistical patterns specific to the training data.

In addition to traditional leaderboards, which often rank SOTA systems based solely on holistic metrics such as aggregate F1-score, \citealp{liu-etal-2021-explainaboard} propose leaderboards that incorporate more fine-grained metrics and offer functionality for direct analysis of misclassifications. This approach allows for a more detailed comparison of system performance, enabling users to directly identify the strengths and weaknesses of specific systems, thereby enhancing the transparency of leaderboards.












\section{Transparency in Benchmark Creation}
\label{sec:transp_bench_creation}

Current relation extraction benchmarks still lack transparency in their creation processes, making it difficult to assert that they generalise well on out-of-distribution data. For instance, we often lack detailed information about the text sources used to create these benchmarks. Transparency in the creation of RE datasets is crucial not only for mitigating potential biases but also for facilitating progress in the field. By better understanding the limitations of existing RE benchmarks we are able to consequently better understand the limitations of systems that make use of these data. We examine the problem of lacking transparency through the lens of two  of the most widely used general-purpose relation extraction benchmarks, namely NYT \citep{10.1007/978-3-642-15939-8_10} and TACRED \citep{zhang-etal-2017-position} datasets. These benchmarks are broadly accepted by the NLP community and continue to be widely used, even in the era of LLMs \citep{huguet-cabot-navigli-2021-rebel-relation,wang-etal-2021-zero,tang-etal-2022-unirel,wang-etal-2022-deepstruct,efeoglu2024retrievalaugmentedgenerationbasedrelationextraction,sainz2024gollie}. Both the NYT and TACRED datasets address the task of sentence-level relation extraction.

\subsection{Analysis of NYT and TACRED Datasets: Transparency and Limitations}

The NYT dataset contains 24 relation types as well as a `None' class and is based on a corpus of New York Times newspaper articles \citep{10.1007/978-3-642-15939-8_10}. As Table \ref{nyt-positive-negative-table} shows, the dataset includes over 266k sentences, with 64\% of the instances belonging to the `None' class.

\begin{table}[!htp]
\centering
\caption{NYT Dataset}
\label{nyt-positive-negative-table}
\scriptsize 
\sisetup{
    detect-all,
    group-digits=true,
    group-separator={,},
    input-ignore={,},
    output-decimal-marker={.},
    table-number-alignment=center,
}
\begin{tabular}{
  @{}
  l
  S[table-format=6.0, table-space-text-post={*}]
  @{}
}
\toprule
\textbf{Type} & {\textbf{Number of Samples}} \\
\midrule
Positive Samples & 96,228 \\
Negative Samples & 170,021 \\
\midrule
\textbf{Total} & 266,249 \\
\bottomrule
\end{tabular}
\end{table}

\begin{figure}[ht]
    \centering
    \includegraphics[width=0.5\textwidth]{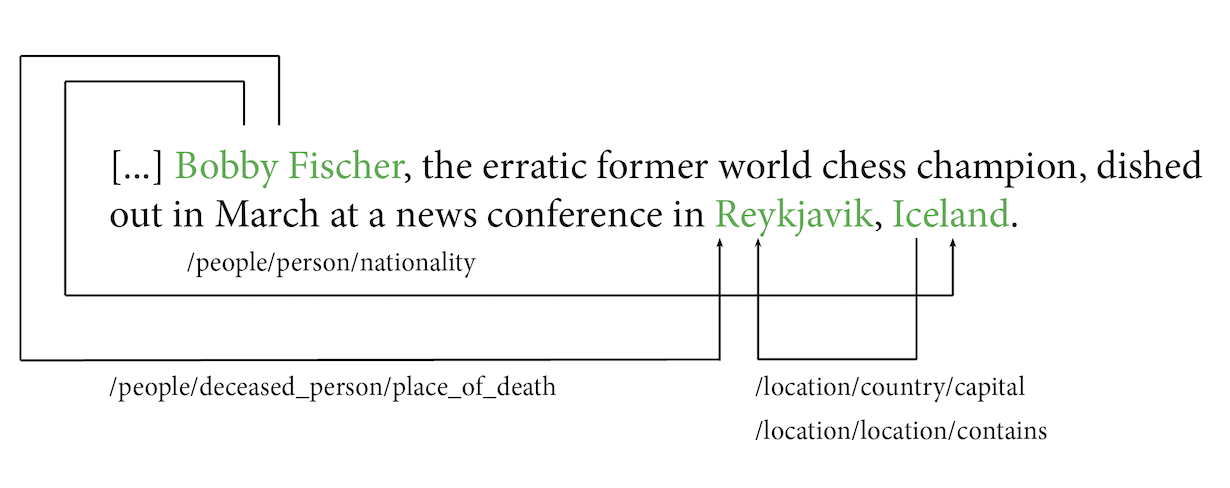}
    \caption{Example from the NYT dataset}
    \label{fig:nyt_example}
\end{figure}

NYT is created through distant supervision, utilising corpus of the New York Times articles \citep{Sandhaus2008Nyt} and using \texttt{Freebase} \citep{10.1145/1376616.1376746} as the external supervision source. Detailed information on the included relation types and the number of instances for each relation can be found in Table \ref{nyt-table} in the Appendix. The NYT dataset is publicly available. The example in Figure \ref{fig:nyt_example} shows one of the instances from the NYT dataset, which illustrates the issues associated with using distant supervision for dataset creation.\footnote{The example in Figure \ref{fig:nyt_example} represents NYT instance with article ID `/m/vinci8/data1/riedel/projects/relation/kb/nyt1/docstore/nyt-2005-2006.backup/1677367.xml.pb'. The NYT dataset can be found at \url{https://github.com/INK-USC/ReQuest}.}

As illustrated by the NYT instance in Figure \ref{fig:nyt_example}, the sentence is labeled as containing the relation `/people/person/nationality' between the head entity \texttt{Bobby Fischer} and the tail entity \texttt{Iceland}. However, this relation is not directly mentioned in text. This issue arises from the distant supervision method used to create the NYT dataset: when named entities are connected by a `nationality' relation in \texttt{Freebase}, it though does not necessarily mean that this relation is explicitly present in the NYT data. Such interpretations can introduce significant biases in relation extraction systems and do not, for instance, reliably demonstrate a system's ability to detect `nationality' relation in general. Such a reasoning pattern can be questioned as valid and would probably be labeled as a hallucination in the era of LLMs. The problem of noise in relation extraction datasets created using distant supervision has been discussed in several works, including \citealp{yaghoobzadeh-etal-2017-noise}.

The TACRED dataset contains 41 relations as well as a `no\_relation' class. TACRED includes over 106k instances, though, as shown in Table \ref{positve-negative-table-tacred}, 80\% of the instances belong to `no\_relation' class, making the dataset highly imbalanced.

\begin{table}[!htp]
\centering
\caption{TACRED Dataset}
\label{positve-negative-table-tacred}
\scriptsize 
\sisetup{
    detect-all,
    group-digits=true,
    group-separator={,},
    input-ignore={,},
    output-decimal-marker={.},
    table-number-alignment=center,
}
\begin{tabular}{
  @{}
  l
  S[table-format=6.0, table-space-text-post={*}]
  @{}
}
\toprule
\textbf{Type} & {\textbf{Number of Samples}} \\
\midrule
Positive Samples & 21,773 \\
Negative Samples & 84,491 \\
\midrule
\textbf{Total} & 106,264 \\
\bottomrule
\end{tabular}
\end{table}
The TACRED dataset is a fully supervised dataset obtained via crowdsourcing, and is based on the TAC KBP\footnote{\url{https://tac.nist.gov/2017/KBP/index.html}} corpus, which includes English newswire and web text. It is distributed under the Linguistic Data Consortium (LDC) license. Detailed information on the included relation types and the number of instances for each relation in TACRED can be found in Table \ref{tacred-table} in the Appendix. The example in Figure \ref{fig:tacred_example} shows one of the instances from the TACRED dataset.\footnote{The example in Figure \ref{fig:tacred_example} originates from the paper describing TACRED \citep{zhang-etal-2017-position}.}

\begin{figure}[ht]
    \centering
    \includegraphics[width=0.5\textwidth]{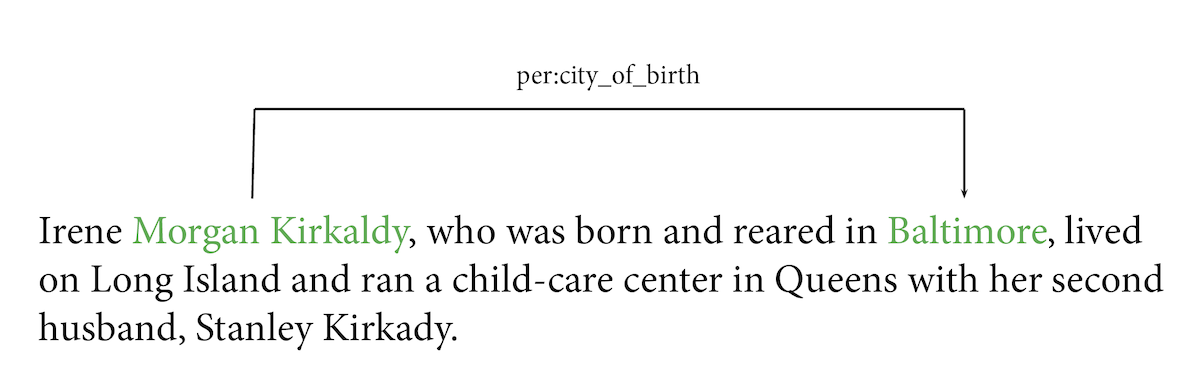}
    \caption{TACRED example}
    \label{fig:tacred_example}
\end{figure}

Compared to the NYT example in Figure \ref{fig:nyt_example}, where the relation type cannot be determined solely from the information provided in the sentence, the TACRED instance in Figure \ref{fig:tacred_example} demonstrates an explicit relation that can be directly extracted from the sentence without requiring additional, potentially biased, reasoning steps. However, the TACRED dataset restricts each sentence to contain only one relation, in contrast to the NYT dataset, which allows each sentence to have multiple labels. The formulation of the relation extraction task like in TACRED can lead to many false negatives \citep{xie-etal-2021-revisiting}. For instance, the example above contains relations like `per:stateorprovinces\_of\_residence', and `per:employee\_of' or `per:spouse', all of which are part of the TACRED list of relations. Thus, restricting each instance to a single relation is far removed from the complexity of real-world text and may significantly mislead the model.

Despite the supplementary material\footnote{\citealp{zhang-etal-2017-position} mention supplementary material to the main paper describing the dataset, though they do not provide the direct link to material. It can be assumed they were referring to information available at \url{https://nlp.stanford.edu/projects/tacred/} or \url{https://tac.nist.gov/2017/KBP/ColdStart/guidelines.html}, but these sources still lack precise details on e.g. the data collection process and the creation of the relations inventory.} provided additionally to the paper \citep{zhang-etal-2017-position}, TACRED still lacks transparency regarding the limitations mentioned above, as well as clarity on how the instances for the dataset were selected from the TAC KBP corpus. A similar lack of transparency exists with the NYT dataset, where the selection process for instances from the NYT corpus \citep{Sandhaus2008Nyt} is not clearly documented. Since access to the TAC KBP corpus is restricted and the selection process from the NYT corpus is unclear, analysing the data included in both datasets and estimating the generalisation capabilities of models trained on these datasets becomes even more challenging.

Additionally, although TACRED instances were manually annotated by crowd workers, unlike the NYT instances, the crowdsourced annotations may still be quite noisy. For instance, \citealp{alt-etal-2020-tacred} demonstrated that over 50\% of the challenging `no\_relation' instances in the development and test sets of TACRED were mislabeled. 

As shown in Tables \ref{nyt-positive-negative-table} and \ref{positve-negative-table-tacred}, both NYT and TACRED are highly imbalanced. While it may appear that the NYT dataset (with 64\% of the instances belonging to the `None' class) is more balanced compared to TACRED (where 80\% of the instances belong to `no\_relation' class), a closer examination on the number of instances for each of the 24 relation types, as detailed in Table \ref{nyt-table} in the Appendix, reveals a different picture. Nearly half of the positive instances in the NYT dataset belong to a single relation type, `/location/location/contains', and six out 24 relations are represented by fewer than 50 instances. For instance, the relation `/people/person/profession' contains only two instances, and `/business/company/industry' has just one instance. 

In addition to providing precise information on the data selection process, it is also important to make the annotation guidelines publicly available if human annotators were involved, or to describe the algorithms used if a dataset was created through distant supervision (e.g., prompts). The publication of annotation guidelines and dataset description in general is among others crucial for clarifying ambiguous relations, whose scope may be understood in multiple ways, such as `per:title' in TACRED or `None' in NYT. The `None' label in NYT could indicate either that none of the 24 specified relations apply — implying another, unspecified relation — or that there is no relation at all between the entities. These are fundamentally different scenarios, and conflating them could lead to significant confusion in the model.

Despite the publicly available TAC KBP guidelines\footnote{\url{https://tac.nist.gov/2014/KBP/ColdStart/guidelines/TAC_KBP_2014_Slot_Descriptions_V1.4.pdf}}, it remains unclear whether this version of annotation guidelines was also provided to the TACRED crowd workers. Furthermore, it is still unclear how the annotators of the TACRED dataset were instructed to handle sentences that contained a relation not listed among the 41 relations, or how they were to deal with sentences containing multiple different relations, as in Figure \ref{fig:tacred_example} above.

To introduce clarity in benchmark creation process, it is therefore crucial to publish not only annotation guidelines but also the instructions provided to the annotators. While \citealp{10.1007/978-3-642-15939-8_10} describes the process of creating the NYT dataset in a relatively detailed way, when they mention the use of human annotators to evaluate a fixed number of extracted relations in a distant supervision scenario, they still do not provide details on how these human annotators were instructed.

\subsection{The Need for Standardised Benchmark Documentation}

The analysis of widely used NYT and TACRED RE benchmarks, along with their available documentation, underscores the persistent issue of lacking exhaustive documentation regarding the creation processes of NLP benchmarks. Proper documentation should be easily discoverable and ideally stored according to accepted standards. Currently, information on NLP benchmarks is dispersed across many resources and often lacks the necessary details to make the benchmark creation process fully transparent which is among others crucial for the analysis of generalisation capabilities of a particular dataset. While sources like \texttt{PaperswithCode}\footnote{\url{https://paperswithcode.com/}} are helpful, they still miss a significant amount of information needed to achieve this goal.

\citealp{10.1145/3458723} addresses the issue of insufficient benchmark transparency and suggests that each new benchmark should be accompanied by a datasheet. The suggested datasheet would include information such as potential sources of noise and errors in the dataset, to enhance transparency and allow for more accurate assessments of the dataset reliability and generalisation capabilities.

The NLP community would greatly benefit from a standardised approach to benchmark documentation, similar to model cards for model reporting \citep{10.1145/3287560.3287596}, but specifically designed for datasets. This is at least as important as model metadata. Model cards, which are essentially files containing metadata with useful information about a model in question, have proven effective, as seen in their implementation at \texttt{HuggingFace}.\footnote{More on model cards at \texttt{HuggingFace} can be found at \url{https://huggingface.co/docs/hub/en/model-cards}} A similar concept for datasets would ensure that critical information about benchmark creation, potential biases, and other relevant details are systematically recorded and easily accessible. While HuggingFace provides dataset cards\footnote{\url{https://huggingface.co/docs/hub/en/datasets-cards}} \citep{park-jeoung-2022-raison} which are a promising step in this direction, most datasets shared via \texttt{HuggingFace} currently have only a fraction of the possible metadata filled out. Moreover, while the ecosystem that \texttt{HuggingFace} provides has undoubtedly contributed significantly to the NLP community, it is essential to acknowledge that, given the open nature of the platform where anyone can upload models and datasets, the reliability of sources, including datasets and their associated metadata, should be approached with caution. Ideally, comprehensive documentation of benchmarks should originate directly from their creators.

A datasheet for benchmarks would ideally include properties such as descriptive and social impact metadata \citep{park-jeoung-2022-raison} including data provenance, data preprocessing details (e.g., filtering approach used to obtain relevant samples), annotation guidelines and other instructions, dataset size, recommended data split information, a list of labels, the specific task being addressed, the method used for creating the benchmark (e.g., human annotation or distant supervision), inter-annotator agreement (if human annotators were involved), and potential sources of noise (e.g., representativeness of the data). In addition to the proposed datasheets for datasets by \citealp{10.1145/3458723}, inspiration can be drawn from dataset templates\footnote{For instance, the \url{https://orkg.org/template/R178304} dataset template contains 22 properties like inter-annotator agreement or data availability.} available in the Open Research Knowledge Graph \citep{10.1007/978-3-030-30760-8_31}. Although these templates are currently used infrequently by benchmark creators, and often only a small fraction of the possible properties are filled out, their wider adoption could significantly enhance benchmark transparency. For instance, a centralised, standardised approach to documenting benchmarks could help establish a more universal system of labels, making it easier to compare benchmarks within a particular domain: e.g., in the case of RE benchmarks, a standardised set of relations could simplify comparisons across different datasets and models.

The way we document the benchmark creation process is becoming increasingly critical in the era of LLMs, especially as we strive to develop Artificial General Intelligence (AGI) systems with human-like reasoning capabilities \citep{Chollet2019OnTM,hendrycks2021measuring}. As we exhaust real-world data, and with the uncertainty about whether data presented as human annotations were truly annotated by humans or generated through LLM prompting, ensuring transparent and thorough documentation is essential for accurately evaluating the systems based on these benchmarks.

\section{Transparency in Leaderboard Performance Evaluation}

The transparency in the benchmark creation process has a direct impact on the ability to adequately evaluate the system trained on a dataset in question and therefore analyse its generalisation capabilities. One of the ways of measuring progress in particular NLP field are leaderboards. Despite the fact that leaderboards push the NLP field forward, they also lack transparency on evaluation process and mostly are limited to the ranking based on holistic metrics such as accuracy or F1-score \citep{liu-etal-2021-explainaboard}. For instance, both TACRED\footnote{\url{https://paperswithcode.com/sota/relation-extraction-on-tacred}} and NYT\footnote{\url{https://paperswithcode.com/sota/relation-extraction-on-nyt}} leaderboards on a widely used platform \texttt{PaperswithCode} rely on F1-score as a holisitic metric to rank the RE models.

Moreover, not only traditional leaderboards lack fine-grained metrics in their ranking approach, but also the papers that report SOTA results follow this trend,  which can lead to an emphasis on achieving top leaderboard positions rather than genuinely addressing the underlying task — a phenomenon known as SOTA-chasing \citep{rodriguez-etal-2021-evaluation}.

Recent papers reporting SOTA-performance on NYT, TACRED,  and TACRED-RE\footnote{TACRED-RE is a revised version of original TACRED, with a subset of challenging development and test set instances relabeled by professional annotators \citep{alt-etal-2020-tacred}.} \citep{huguet-cabot-navigli-2021-rebel-relation,wang-etal-2022-deepstruct,tang-etal-2022-unirel,efeoglu2024retrievalaugmentedgenerationbasedrelationextraction,sainz2024gollie,orlando-etal-2024-relik} report only aggregate metrics such as micro or macro f1-score, recall, and precision. Consequently, these evaluations lack more fine-grained, class-based metrics, which are crucial for the analysis of RE systems dealing with a large number of relations such as the 42 labels in TACRED and the 25 labels in NYT. In the context of imbalanced datasets like TACRED and NYT, a system may achieve high overall metrics by always predicting a `no\_relation' class. However, this outcome does not indicate that the system has indeed effectively learned to solve the relation extraction task across a diverse set of over 20 labels. Notably, even the original papers presenting TACRED \citep{zhang-etal-2017-position}, TACRED-RE \citep{alt-etal-2020-tacred}, NYT \citep{10.1007/978-3-642-15939-8_10} do not contain fine-grained, class-based metrics. Given the significant class imbalance reflected in the Tables \ref{nyt-positive-negative-table} and \ref{positve-negative-table-tacred}, as well as the fact that many relations in both TACRED and NYT are represented by only a few instances, such as `/people/person/profession' in NYT (see Table \ref{nyt-table} in the Appendix), which contains only two instances, reporting class-based metrics is essential for adequately assessing the capabilities of a particular system to solve the RE task. Without detailed performance reports, it is difficult to determine whether a new SOTA system generalises well or simply creates the illusion of improvement through SOTA-chasing.

Benchmarks such as NYT, TACRED, and TACRED-RE lack standardised guidelines for reporting results, leading to inconsistencies across publications that report SOTA results on these benchmarks \citep{dehghani2021benchmarklottery}. This lack of agreement can cause discrepancies in leaderboard rankings. For example, there is no consensus on which aggregated score should be used on platforms like \texttt{PaperswithCode}. The current top-performing model on the TACRED benchmark \citep{efeoglu2024retrievalaugmentedgenerationbasedrelationextraction} reports the micro-F1 score, which is also used for ranking. In contrast, the current second \citep{wang-etal-2022-deepstruct} and third \citep{huang-etal-2022-unified} top-ranked models on TACRED report the macro-F1 score, which is also utilised for their ranking on \texttt{PaperswithCode}. This inconsistency in evaluation metrics raises concerns about the reliability of the leaderboard rankings.

Additionally, there is no overlap between the top-performing models listed on \texttt{PaperswithCode} leaderboards for NYT and TACRED, meaning that all top-performing models for TACRED are different from those for NYT. This further complicates the analysis of these models' generalisation capabilities and makes it difficult to assess model ranking consistency across RE benchmarks. Focusing exclusively on achieving high performance on a single benchmark, without considering results across multiple benchmarks, can result in models that are overly specialised for specific benchmarks. This, however, does not necessarily indicate meaningful progress in addressing a particular NLP task \citep{dehghani2021benchmarklottery}, such as relation extraction.

Papers reporting SOTA results on RE, including the original TACRED \citep{zhang-etal-2017-position}, TACRED-RE \citep{alt-etal-2020-tacred}, NYT \citep{10.1007/978-3-642-15939-8_10}, often do not provide information on whether the issue of class imbalance was addressed. Such details should be included in system description papers, particularly when reporting new SOTA results. For instance, the authors of the Biographical RE dataset \citep{10.1145/3477495.3531742} tackled the problem of large class imbalance by removing some of majority class relations, thereby equalising them with the sum of all other relations.

Model performance ceiling \citep{alt-etal-2020-tacred} may be caused by the presence of noisy data, which can limit the potential for improvement by new RE methods. As discussed in Section \ref{sec:transp_bench_creation}, this noise can originate from both distantly-supervised datasets, such as NYT, and fully-supervised crowdsourced datasets, such as TACRED. Additionally, the way the task is formulated, whether as a single-label (TACRED) or multi-label (NYT) classification task, can contribute to performance limitations. For example, an RE model might make a correct prediction, but due to the task being framed as a single-label classification problem — despite real-world instances potentially containing multiple relations — this could lead to misclassification. Such factors should be considered when reporting new SOTA results. Moreover, in the era of LLMs, it is possible that multiple outputs generated by an LLM for an RE task could be correct \citep{hendrycks2021measuring}, a nuance that is not captured by holistic metrics like aggregate F1-score.




Due to the mentioned limitations of traditional leaderboards such as the ones utilised on the \texttt{PaperswithCode} platform, \citealp{liu-etal-2021-explainaboard} suggest an ExplainaBoard interactive tool that provides both holistic and fine-grained metrics as well as functionality for direct analysis of misclassifications. Such an extension of traditional leaderboards enables the direct detection of strengths and weaknesses of a particular system, as well as of a benchmark, thereby enhancing the ability to assess the generalisation capabilities of systems, such as those used for relation extraction.

Moreover, evaluating model performance on a test set drawn from the same distribution as the training set does not necessarily demonstrate a model's ability to solve an underlying task \citep{linzen-2020-accelerate}, such as relation extraction. To address this issue, \citealp{linzen-2020-accelerate} propose Generalisation Leaderboards, which would evaluate systems on test sets derived from different distributions than the training set. For instance, it would be valuable to assess a system fine-tuned on TACRED data for its ability to extract the same subset of relations present in the NYT dataset, as strong performance on one dataset does not necessarily indicate robust generalisation capabilities. Additionally, techniques such as adversarial attacks \citep{wu-etal-2019-errudite,gardner-etal-2020-evaluating,goel-etal-2021-robustness} can further test the true capabilities of RE systems by exposing their vulnerabilities and resilience to challenging scenarios.




\section{Conclusion and Future Work}

In this work, we have highlighted several limitations in the benchmark documentation and use of traditional leaderboards, particularly those employed for the relation extraction task. Limitations in benchmark documentation include the absence of comprehensive descriptive metadata, such as the source of the data or details regarding inter-annotator agreement, as well as an absence of clear articulation of the dataset's inherent limitations, such as large class imbalances and potential noise. Furthermore, there is often insufficient discussion on methods to mitigate these issues.

Evaluating systems based on these RE benchmarks inherently necessitates addressing the problems associated with insufficient documentation of the benchmarks. For instance, traditional leaderboards, such as those on \texttt{PaperswithCode}, that play a significant role in advancing NLP, typically rely on holistic metrics like F1-score. However, these metrics fail to capture the complexity of the relation extraction task, especially in scenarios involving a large number of labels and highly imbalanced datasets, such as TACRED, where most instances belong to a `no\_relation' class. Additionally, papers reporting new SOTA results on RE benchmarks like NYT and TACRED often focus exclusively on aggregate metrics, neglecting class-based metrics, which obscures the nuanced performance of models across different relation types.

This paper does not intend to undermine the significance and value of existing benchmarks such as TACRED or NYT, which are crucial for the evaluation of models in the field, as well as the development of new SOTA approaches. Instead, given the evolving perspective on data used in training deep learning models, our objective is to propose avenues for improving the documentation of benchmark creation processes, which would in turn help to better assess the generalisation capabilities of RE models. Additionally, we also aim to motivate the adoption of more rigorous evaluation practices, encouraging researchers to move beyond the limited scope of only reporting metrics such as aggregate F1-score, precision, and recall. This is particularly important in NLP tasks such as relation extraction, where the complexity is exacerbated by the presence of a large number of relations.

It is also crucial to recognise that high performance on a specific RE benchmark, such as TACRED, TACRED-RE, or NYT, reflects only a model's ability to handle a subset of all possible relations. Furthermore, even if a system performs well on a given subset of relations, it may struggle significantly when extracting the same relations from out-of-distribution data.

Our focus should not solely be on the development of new approaches, but also on critically analysing our systems and recognising the limitations of the data used for their evaluation. This critical perspective is essential for advancing the field and ensuring that our models are robust and generalisable.

Finally, this work serves as a position paper that highlights several issues in the creation of RE benchmarks and the practices surrounding leaderboard evaluations. We acknowledge the limitations of this work, particularly the lack of extensive quantitative evidence. In our future research, we aim to conduct a comprehensive cross-dataset evaluation of RE systems on the benchmarks discussed. Such an evaluation will provide empirical support for the concerns raised and offer a more reliable assessment of the generalisation capabilities of current RE systems.

\bibliography{anthology,custom}

\appendix
\clearpage
\section{Dataset Statistics: Class Distribution}
\label{sec:appendix}

\begin{table}[!htp]
\centering
\caption{TACRED Dataset}
\label{tacred-table}
\scriptsize 
\sisetup{
    detect-all,
    group-digits=true,
    group-separator={,},
    input-ignore={,},
    output-decimal-marker={.},
    table-number-alignment=center,
}
\NewDocumentCommand{\B}{}{\fontseries{b}\selectfont}
\begin{tabular}{
  @{}
  l
  S[table-format=7.0, table-space-text-post={*}]
  @{}
}
\toprule
\textbf{Relation} & {\textbf{\# of Samples}} \\
\midrule
no\_relation & 84,491 \\
per:title & 3,862 \\
org:top\_members/employees & 2,770 \\
per:employee\_of & 2,163 \\
org:alternate\_names & 1,359 \\
per:age & 833 \\
per:countries\_of\_residence & 819 \\
org:country\_of\_headquarters & 753 \\
per:cities\_of\_residence & 742 \\
per:origin & 667 \\
org:city\_of\_headquarters & 573 \\
per:stateorprovinces\_of\_residence & 484 \\
per:spouse & 483 \\
org:subsidiaries & 453 \\
org:parents & 444 \\
per:date\_of\_death & 394 \\
org:stateorprovince\_of\_headquarters & 350 \\
per:children & 347 \\
per:cause\_of\_death & 337 \\
per:other\_family & 319 \\
per:parents & 296 \\
org:members & 286 \\
per:charges & 280 \\
org:founded\_by & 268 \\
per:siblings & 250 \\
per:schools\_attended & 229 \\
per:city\_of\_death & 227 \\
org:website & 223 \\
org:member\_of & 171 \\
org:founded & 166 \\
per:religion & 153 \\
per:alternate\_names & 153 \\
org:shareholders & 144 \\
org:political/religious\_affiliation & 125 \\
org:number\_of\_employees/members & 121 \\
per:stateorprovince\_of\_death & 104 \\
per:date\_of\_birth & 103 \\
per:city\_of\_birth & 103 \\
per:stateorprovince\_of\_birth & 72 \\
per:country\_of\_death & 61 \\
per:country\_of\_birth & 53 \\
org:dissolved & 33 \\
\midrule
\textbf{Positive Samples} & 21,773 \\
\textbf{Negative Samples} & 84,491 \\
\textbf{Total} & 106,264  \\
\bottomrule
\end{tabular}
\end{table}

\begin{table}[!htp]
\centering
\caption{NYT Dataset}
\label{nyt-table}
\scriptsize 
\sisetup{
    detect-all,
    group-digits=true,
    group-separator={,},
    input-ignore={,},
    output-decimal-marker={.},
    table-number-alignment=center,
}
\NewDocumentCommand{\B}{}{\fontseries{b}\selectfont}
\begin{tabular}{
  @{}
  l
  S[table-format=7.0, table-space-text-post={*}]
  @{}
}
\toprule
\textbf{Relation} & {\textbf{\# of Samples}} \\
\midrule
None & 170,021 \\
/location/location/contains & 44,490 \\
/location/country/capital & 7,267 \\
/people/person/nationality & 7,244 \\
/people/person/place\_lived & 7,015 \\
/location/administrative\_division/country & 5,951 \\
/location/country/administrative\_divisions & 5,851 \\
/business/person/company & 5,421 \\
/location/neighborhood/neighborhood\_of & 5,082 \\
/people/person/place\_of\_birth & 3,133 \\
/people/deceased\_person/place\_of\_death & 1,914 \\
/business/company/founders & 767 \\
/people/person/children & 487 \\
/business/company/place\_founded & 414 \\
/business/company/major\_shareholders & 282 \\
/business/company\_shareholder/major\_shareholders & 282 \\
/sports/sports\_team\_location/teams & 218 \\
/sports/sports\_team/location & 218 \\
/people/person/religion & 67 \\
/business/company/advisors & 45 \\
/people/ethnicity/geographic\_distribution & 33 \\
/people/ethnicity/people & 21 \\
/people/person/ethnicity & 21 \\
/people/person/profession & 2 \\
/business/company/industry & 1 \\
\midrule
\textbf{Positive Samples} & 96,228 \\
\textbf{Negative Samples} & 170,021 \\
\textbf{Total} & 266,249 \\
\bottomrule
\end{tabular}
\end{table}

\end{document}